\documentclass[journal]{IEEEtran}
\usepackage{siunitx}
\usepackage{nicefrac}
\usepackage{cite}
\ifCLASSINFOpdf
  \usepackage[pdftex]{graphicx}
  \graphicspath{{../pdf/}{../jpeg/}}
  \DeclareGraphicsExtensions{.pdf,.jpeg,.png}
\else
\fi
\usepackage{amsmath}
\usepackage{amssymb}
\usepackage{url}
\begin{document}

% this removes the url from biblio: Eurico Covas
\bstctlcite{IEEEexample:BSTcontrol}
% this removes the url from biblio: Eurico Covas

\title{Optimal Neural Network Feature Selection for Spatial-Temporal Forecasting}

% \author{Michael~Shell,~\IEEEmembership{Member,~IEEE,}
        % John~Doe,~\IEEEmembership{Fellow,~OSA,}
        % and~Jane~Doe,~\IEEEmembership{Life~Fellow,~IEEE}% <-this % stops a space
% \thanks{M. Shell was with the Department
% of Electrical and Computer Engineering, Georgia Institute of Technology, Atlanta,
% GA, 30332 USA e-mail: (see http://www.michaelshell.org/contact.html).}% <-this % stops a space
% \thanks{J. Doe and J. Doe are with Anonymous University.}% <-this % stops a space
% \thanks{Manuscript received April 19, 2005; revised August 26, 2015.}}

\author{Eurico~Covas and
	    Emmanouil~Benetos,~\IEEEmembership{Member, IEEE}% <-this % stops a space
\thanks{E. Covas is with the CITEUC, Geophysical and Astronomical Observatory, University of Coimbra, 3040-004, Coimbra, Portugal, and
the School of Electronic Engineering and Computer Science, Queen Mary University of London, Mile End Road, London E1 4NS, U.K.,  e-mail: eurico.covas@mail.com}% <-this % stops a space
\thanks{E. Benetos is with the School of Electronic Engineering and Computer Science, Queen Mary University of London, Mile End Road, London E1 4NS, U.K.,
e-mail: emmanouil.benetos@qmul.ac.uk}% <-this % stops a space}\thanks{Manuscript received December 31, 2017; revised December 31, 2017.}
\thanks{Manuscript received x, 2018; revised xx, 2018.}}


\markboth{}%
{Covas \MakeLowercase{\textit{et al.}}: Optimal Neural Network Feature Selection for Spatial-Temporal Forecasting}



\maketitle

\begin{abstract}
In this paper, we show empirical evidence on how to construct the optimal feature selection or input representation used by  
the input layer of a feedforward neural 
network for the propose of forecasting spatial-temporal signals. The approach is based on results from dynamical systems theory, namely 
the non-linear embedding theorems. We demonstrate it for a variety of spatial-temporal signals, with one spatial and one temporal 
dimensions, and show that the optimal input layer representation consists of a grid, with spatial/temporal lags determined by 
the minimum of the mutual information of the spatial/temporal signals and the number of points taken in space/time decided by the 
embedding dimension of the signal. We present evidence of this proposal by running a Monte Carlo simulation of several combinations of 
input layer feature designs and show that the one predicted by the non-linear embedding theorems seems to be optimal or close of 
optimal. In total we show evidence in four unrelated systems: a series of coupled H\'{e}non maps; a series of couple Ordinary 
Differential Equations (Lorenz-96) phenomenologically modelling atmospheric dynamics; the Kuramoto-Sivashinsky equation, a partial
differential equation used in
studies of instabilities in laminar flame fronts and finally real physical 
data from sunspot areas in the Sun (in latitude and time) from 1874 to 2015.
\end{abstract}


% \fi
\IEEEpeerreviewmaketitle

\section{Introduction}
\IEEEPARstart{G}{iven} a physical data set, one of the most important questions one can pose is: ``Can we predict the future?'' This 
question can be put forward irrespectively of the fact that we may already have some insight or even be certain on what the exact model 
behind some or all the observed variables is. For example, for chaotic dynamical systems \cite{0813340853,2004icti.book.....M}, we may 
even have the underlying dynamics but still find it hard to predict the future, given that chaotic systems have exponential sensitivity 
to initial conditions. The more chaotic a system is (as measured by the positiveness of their largest Lyapunov exponents 
\cite{1985PhyD...16..285W,1994PhLA..185...77K}) the harder it gets to predict the future, even within very short time horizons. In the 
limit case of a random system, it is not possible to predict the future at all, although one can opine on certain future 
statistics\cite{9780486693873}. For the case of weakly chaotic systems, there is an extensive literature on forecasting methods ranging 
from linear approximations\cite{1451722}; truncated functional expansion 
series\cite{Powell:1987:RBF:48424.48433,Broomhead1988MultivariableFI}; non-linear embeddings \cite{PhysRevLett.59.845}; auto-regression 
methods\cite{0130607746}; hidden Markov models \cite{1165342} to state-of-the-art neural networks and deep learning methodologies 
\cite{LANGKVIST201411} and many others, too long to list here.

Most literature on forecasting chaotic signals is dedicated to a single time series, or treat a collection of related time series as a 
non-extended set, i.e. a multivariate set of discrete variables as opposed to a spatially continuous series. 
For forecasting spatial-temporal chaos we 
refer the reader to 
\cite{doi:10.1063/1.165894,PhysRevE.51.R2709,PhysRevLett.85.2300,ORSTAVIK1998145,
Parlitz2000NonlinearPO,2000PhRvL..84.1890P,Covas,Xia2006APF,ENV:ENV2266,covas2016,ENV:ENV2456} and 
references therein. Even rarer are attempts to forecast spatial-temporal chaos using neural networks and deep learning methodologies 
\cite{covaspeixinhojoao,2017arXiv170805094M,2017arXiv171100636M,2017arXiv171110566R,2017arXiv171110561R,2017arXiv171009668L, 
2017arXiv171205293C, ghaderi2017deepforecast,2017Chaos..27d1102L,
2018JCoPh.357..125R,2018arXiv180106637R}, although this field of research is clearly growing at the moment\footnote{
There is also a new emerging field of research on solving PDEs (therefore implicitly predicting a spatial-temporal
evolution) using deep learning -- see \cite{2017arXiv170905963B,2017arXiv170807469S,2017arXiv170604702E} and references therein.
Furthermore, notice that in this article we are concerned with the full space-time prediction, 
as opposed to the ongoing research on pattern recognition in moving 
images (2D and 3D), which attempt to pick particular features (e.g. car, pedestrian, bicycle, person, etc.) and to forecast where 
those features will be in subsequent images within a particular moving sequence -- see \cite{article_motion} and reference therein.
}.
 Nonetheless, this area of 
research is of importance, as most physical systems are spatially extended, e.g. the atmospheric system driving the Earth's weather
\cite{9780521857291}; the solar dynamo driving the Sun's sunspots \cite{9780198512905}; and the influence of sunspots on the Earth's 
magnetic field via the solar wind, coronal mass ejections and solar flares -- the so-called space weather 
\cite{1851RSPT..141..123S, 1852RSPT..142..103S, 1979P&SS...27.1001S,
1983SoPh...89..195E,1965P&SS...13....9P,2000AdSpR..26...27W,2003A&AT...22..861B,2005GeoRL..3221106S,
2005SpWea...3.8C01K,2006GMS...165..367T,2006GeoRL..3318101H, 2009SunGe...4...55C,2011SpWea...9.6001C,
2013EGUGA..1510865W}, which may have real economic implications \cite{2015SpWea..13..524S}.
Nonetheless its importance, forecasting spatial-temporal chaos is difficult.
The reasons are many, but mainly: first, the geometric dimension of the 
attractor \cite{1985PhLA..107..101G} -- usually quite large, the so-called curse of dimensionality \cite{9780486428093}; 
and second how to choose the variables to use for forecasting, i.e., is there 
enough information on the same point back in time to derive the future of that particular point, or do spatial correlations and spatial 
propagation affect it in a way that one must take into account some spatial and temporal neighbours set to forecast the future.
If this is the case, can that set of points be defined and how can it be constructed? 
It is this last question that we investigate in this article, in the 
particular context of spatial-temporal forecasting using neural networks. 

Feature extraction and the design of the input
representation of the 
input layer for a neural network is considered to be an art form, relying mostly on trial and error and domain knowledge (see \cite{5771385}
for examples and references). For 
forecasting of time series, a simple approach consists of designing the input layer as a vector of previous data using a time delay, the 
time delay neural network method \cite{Waibel:1990:PRU:108235.108263, luk2000study, 857903, Frank2001, OH2002249, 1009-1963-12-6-304, 
inputlayer}. For spatial-temporal series, one can generalize it to include temporal and spatial delays 
\cite{covas2016,covaspeixinhojoao}. This is where the connection to dynamical systems can be useful. 

In 1981, Takens established the 
theoretical background \cite{1981LNM...898..366T} in his embedding theorem for a mathematical method to reconstruct the dynamics 
of the underlying attractor of a chaotic dynamical system from a time ordered sequence of data observations. Notice the reconstruction 
conserves the properties of the original dynamical system up to a diffeomorphism. 
Further developments established a series of theorems \cite{key1503303m, 1981LNM...898..230M, 1991JSP....65..579S} that provided the basis for a non-linear embedding and forecasting on the original variables. The 
theorems and related articles propose to use a time delay approach with the time lag based on the first minima of the 
{\em mutual information}\footnote{
Notice that another non-linear dynamical systems technique exists to calculate this time delay, the zero of the autocorrelation
function \cite{opac-b1092652,abarbanel1997analysis}, but essentially these two approaches are after the same objective, i.e.
to select uncorrelated 
 variables as much as possible for optimal reconstruction embedding. So, in this article, we focus only on the first minima
 of the mutual information for simplicity of analysis.
} -- see 
\cite{Fraser86, abarbanel1997analysis, opac-b1092652} -- and to choose the number of points to include using  the method of {\em false 
nearest neighbours} detection suggested by \cite{1992PhRvA..45.3403K} and reported in detail in \cite{1992PhRvA..45.7058M, 
1993RvMP...65.1331A, 1996PhT....49k..86A, abarbanel1997analysis}. 

Some authors discuss the use of either the mutual information and/or embedding dimension as a constraint on feature representation
\cite{annunziato, Gkana201579, Zachilas2015, Sun2010109, HUANG20108590, 298224, Frank2001, 1998GeoRL..25..457K, 
BUHAMRA2003805, chandra2012cooperative, DBLP:journals/corr/MaslennikovaB14, sauter2010spatio, JiangS11, inputlayer, 1997IJMPC...8.1345K, 
1009-1963-12-6-304, Verdes2000, 1996SoPh..168..423F, 2007AdG....10...67L, Chandra:2012:CCE:2181341.2181747, raios, 
maass2003mathematical,0305-4470-28-12-012, 1995ApJ...444..916C, 1998GeoRL..25..457K, Zachilas2015, Chandra:2012:CCE:2181341.2181747, 
Gkana201579, raios}. Others \cite{Simon:2007:HDS:1230147.1230294} attempted to generalize the mutual 
information approach to higher dimensions but do not actually connect it to the problem of spatial-temporal forecasting using neural networks. 
There are also authors \cite{articleRagulskis} that try to use neural networks to determine the optimal embedding and time 
delay for the purpose of local reconstruction of states with a view to forecast (the opposite of what we try to empirically demonstrate 
here). Fig.\ 6 in \cite{1555956} shows how the forecasting error for a pure time series prediction changes with the delay and the number of 
time delay points used as an input -- they use a reinforcement learning based dimension and delay estimator to derive the best dimension 
and delay, but do not seem to show that it is the dynamical systems' derived values that are indeed optimal for forecasting neither
they show any extension to spatial-temporal signals as we demonstrate in this article.
Other authors \cite{Xia2006APF} try to use Support Vector Machines (SVMs) to forecast spatial-temporal signals and use delays and 
embedding approaches to define the state vectors. In fact, Parlitz and Merkwirth \cite{Parlitz2000NonlinearPO} 
%in their ESANN (European Symposium on Artificial Neural Networks) 2000 meeting proceedings' article 
mention
 that local reconstruction of states ``\ldots may also serve as
a starting point for deriving local mathematical models in terms of polynomials,
radial basis functions or neural networks\ldots''. Here we attempt to show empirical evidence that this is not just
a starting point, but the optimal neural network input feature selection. 

We also emphasise that,
 as far as we are aware, all the references on neural network forecasting of spatial-temporal dynamics that use the 
embedding theorems and the related mutual information and the false nearest neighbours methods seem not to justify its use, i.e., 
the approach is explained, even suggested to be optimal, but neither proven theoretically or empirically. Here we attempt 
to provide an empirical evidence for this optimality. 
Using this theoretical framework, we propose that this non-linear 
embedding method, using the training data alone without reference to the forecasting model, 
can be used to indicate the best way to construct the feature representation for 
the input layer of a neural network used for forecasting both in space and time.
In order to support this proposal, we, in this article, show empirical evidence for an optimal feature selection 
 for four particular cases of two-dimensional
spatial-temporal data series $s^n_m$, where by two-dimensional we mean a scalar field that can be
defined by a $N \times M$ matrix with components $s^n_m \in \mathbb{R}$. Furthermore, notice the primary goal is not to demonstrate 
the ability to forecast, which has already been done by several authors in the literature above, but rather that there is no need to calibrate
the neural network feature selection specification by the ``dark art'' of trial and error.
 
The article is divided as follows. In section \ref{modelsection} we explain our forecasting model, in section \ref{conjecturesection} 
we describe our proposal, in section \ref{resultssection} we show our results supporting this proposal and finally in section
\ref{conclusionsection} we make our concluding remarks.

\section{Model}
\label{modelsection}

\begin{figure*}[!htb]
\resizebox{\hsize}{!}{\includegraphics{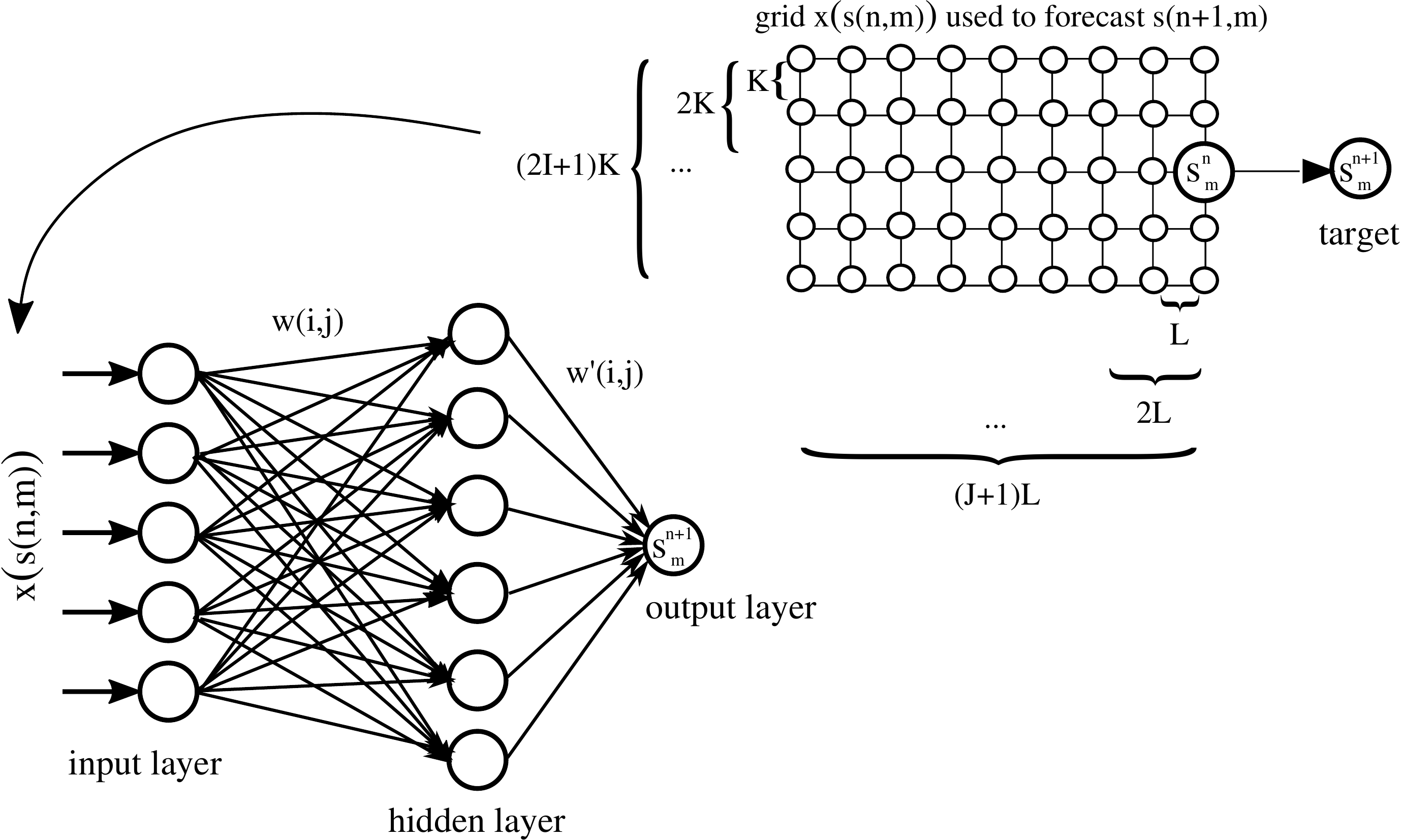}}
\caption{Neural network architecture for forecasting spatial-temporal signals. 
The neural network is made of an input layer, one or more hidden layer(s) and one output layer.
In this article, for simplicity, we use only one hidden layer and the output layer is made of a single neuron. 
Each input pattern $x(i)$ is sent to the
input layer, then each of the hidden neurons' values is calculated from the sum of the product of the weights by the inputs $\sum w(i,j) x(i)$
and passed via the non-linear activation function. Then the output is calculated by the product of the second set of
weights times the hidden node values $\sum w'(i,j) y(i)$ again passed to another (or the same) activation function.
Each input pattern $x(i)$ is actually a matrix constructed using an embedding space of
spatial and temporal delays, calculated from the actual physical spatial-temporal data values $s(n,m)$. After many randomly chosen input patterns
are passed via the neural network, the weights hopefully converge to an optimal training value. One can then forecast using the last
time slices of the training set, and compare against the test set, the real future data set.}
\label{architecture}
\end{figure*}

The neural network architecture we chose to demonstrate our proposal is a form of the basic feedforward neural network,
sometimes called the time-delayed neural network \cite{Waibel:1990:PRU:108235.108263}, trained using the so-called back-propagation
algorithm \cite{10.1007/BFb0006203, 1986Natur.323..533R, 58337, Lecun98gradient-basedlearning}. 
We focus on spatial-temporal series, so we have extended the usual time-delayed neural network
to be a time and space delayed network. The overall feature representation of the network is depicted in detail in 
Fig.\ \ref{architecture}. Notice we chose to use feedforward neural networks rather than more complex neural networks such
as recurrent neural networks\cite{COGS:COGS203}, since
feedforward ones are simpler to design; are capable of being used for forecasting of even complex chaotic signals; are guaranteed
to converge, at least, to a local minima; and are easier to interpret.

% to edit do the following 3 commands
% inkscape architecture.svg (then print to architecture.pdf)
% pdf270 --suffix 'turned' --batch architecture.pdf
% pdfcrop architecture-turned.pdf

Under this input representation, we use the ideas proposed in \cite{Parlitz2000NonlinearPO, 2000PhRvL..84.1890P} to construct a grid of input values
which are then fed to the neural network to produce a single output, the future state. Formally,
let $n=1,...,N$ and $m=1,...,M$.
Consider a spatial-temporal data series ${\bf s}$ which can be
defined by a $N\times M$ matrix with components
$s^n_m \in \mathbb{R}$. To these components, we will call {\em states} of the spatial-temporal series.
Consider a number $2I\in \mathbb{N}$ of neighbours in space of a given
$s^n_m$ and a number $J\in \mathbb{N}$ of temporal past neighbours relative to $s^n_m$ (see Fig.\ \ref{architecture} for details).
For each $s^n_m$, we define the input (feature) vector ${\bf x} (s^n_m)$ with components given by
$s^n_m$, its $2I$ spatial neighbours and its $J$ past temporal
neighbours, and with $K$ and $L$ being the spatial and temporal lags:
\begin{IEEEeqnarray}{rCl}
\label{embedding}
{\bf x}(s^n_m)&=&\{s^n_{m-I K }, ...,s^n_m, ..., s^n_{m+I K},\\
&& {}s^{n-L}_{m-I K},..., s^{n-L}_{m},..., s^{n-L}_{m+I K},
\ldots \nonumber\\ 
&& {}\ldots,s^{n-J L}_{m-I K},..., s^{n-J L}_{m},..., s^{n-J L}_{m+I K} \}
\nonumber
\end{IEEEeqnarray}

So, the input is a
$(2 I+1)(J+1)$ vector ${\bf x}(s^n_m)$ and the target (output) to train the network is the value $s^{n+1}_{m}$.
 We train the network using stochastic gradient
back-propagation by running a stochastic batch
where we randomly sample pairs of inputs and outputs from the training set: ${\bf x}(s^n_m)$ and $s^{n+1}_{m}$, respectively. Then at test time
we chose inputs ${\bf x}(s^n_m)$, such that $n=N_{train}$, $N_{train}$ being the number of temporal slices on the training set.
As for the remaining architecture, we use one hidden layer with $N_h$ nodes.
Regarding the back-propagation hyperparameters, we included an adaptive learning rate $\eta_n=\eta/(1+n/10000)$, 
where the hyperparameter $\eta$ is the initial learning rate and $\eta_n$ is the learning rate used at time step
$n$. We included a momentum $\alpha$ for faster convergence.
 A further hyperparameter is the choice of the activation function (see \cite{9780262527019}), we use either a ReLu (rectified linear unit)
 or a logistic sigmoid function depending on the
 test case we are working with.
We also normalize the data before passing it through the neural network, in most cases we scale it in linear fashion
$x \to \alpha_{nor} + x/\beta_{nor}$, and in the case of real physical data as we will see later, 
we scale it in logarithmic fashion it by $x \to \alpha_{nor}+\nicefrac{\ln(1+x)}{\beta_{nor}}$, where $x$ is the initial data,
and $\alpha_{nor}$ and $\beta_{nor}$ are the arbitrary shift and scaling constants, respectively. For the weight (and bias)
initialization we chose random numbers with a constant distribution between $[0,1]$ and shifted by $\alpha_{rng}$ and scaled
by $\beta_{rng}$. The final hyperparameter is the number of epochs taken on the stochastic gradient descent
which we denote by $N_{\textnormal{steps}}$.  
All of these hyperparameters are calibrated and fixed
before we do any simulations with respect to the parameters $I$, $J$, $K$, $L$, which are auto-calibrated by the above mentioned
methods derived from dynamical systems theory. In this sense, $I$, $J$, $K$, $L$ are not hyperparameters of the neural network. We use the standard
loss function $\mathcal{L}=\left( s^{n+1}_{m} - \hat{s}^{n+1}_{m}\right)^2$ for a prediction $\hat{s}^{n+1}_{m}$ 
centred around $s^{n}_{m}$ using as input
the feature vector ${\bf x}(s^n_m)$ of total dimension  $(2 I+1)(J+1)$.

\section{Proposal}
\label{conjecturesection}

\begin{figure}[!htb]
\resizebox{\hsize}{!}{\includegraphics{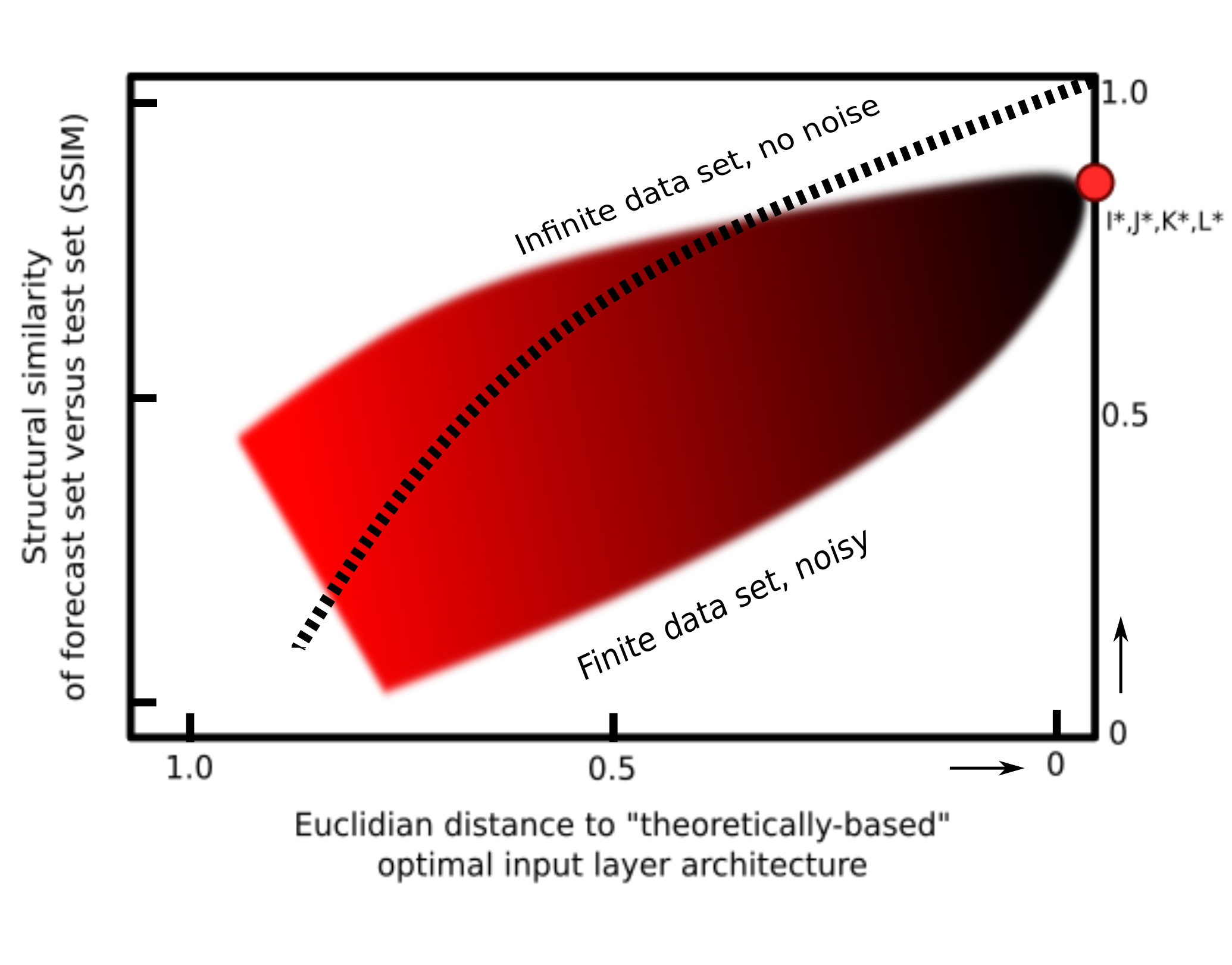}}
\caption{Our main proposal. For a infinite noiseless training set, the SSIM approaches $\textnormal{SSIM} \to 1$. For real data sets,
there is a dispersion of the SSIM versus some reasonable metric constructed to
 represent the distance between any feature selection (e.g.\ $d_e = \sqrt{(I-I^*)^2+(J-J^*)^2+(K-K^*)^2+(L-L^*)^2}$).
}
\label{conjecture}
\end{figure}

Once we do a forecast, we then compare the goodness of fit by first visual inspection and second by numerically calculating the so-called
structural similarity $\textnormal{SSIM}(x,y)$ which has been proposed by \cite{Wang04imagequality} and used already in the context of spatial-temporal forecasting in \cite{covas2016, covaspeixinhojoao}. It has also been used in the context of deep learning used for enhancing
resolution on two dimensional images \cite{2015arXiv150100092D} and restoring missing data in images \cite{2018arXiv180208369Z}. For details on the SSIM measure see \cite{Wang04imagequality,2009ISPM...26...98W, 2012ITIP...21.1488B}.
The SSIM index is a metric quantity used to calculate the perceived quality of digital images and videos.  It
allows two images to be compared and provides a value of their similarity - a value of $\textnormal{SSIM}=1$ corresponds to the case of two
perfectly identical images. We use it by calculating the $\textnormal{SSIM}(x,y)$ between the entire test set and the forecast set,
since these can be interpreted as images (one spatial dimension/one temporal dimension).

Here we propose that the optimal time delay/spatial delays ($L$ and $K$, respectively) must be the ones based on the first minima of the mutual 
information \cite{Fraser86, abarbanel1997analysis, opac-b1092652} and that the optimal number of temporal/spatial points to use ($J$ and $I$, 
respectively) must be the ones based on the method of false nearest neighbours detection \cite{1992PhRvA..45.3403K, 
1992PhRvA..45.7058M, 1993RvMP...65.1331A, 1996PhT....49k..86A, abarbanel1997analysis}. The mutual information is calculated by taking a $s^{i}$, a 
one-dimensional data set, and $s^{i+L}$, the related $L$-lagged data set. Given a measurement $s^i$, the amount of information $I(L)$ is the number 
of bits on $s^{i+L}$, on average, that can be predicted. We then average over space and take the first minimum of $\langle I(L) \rangle$, or, in the 
absense of a clear minimum, take the $L$ temporal lag for which the $\langle I(L) \rangle$ drops significantly and starts to plateau. This calculates 
$L^*$, the optimal time delay. Conversely, we calculate $K^*$ by calculating the spatial lag $K$ for which we obtain the first minima of the 
time-averaged mutual information $\langle I(K) \rangle$. Once the optimal spatial and temporal lags $K^*$ and $L^*$ are calculated, we calibrate 
the minimum embedding dimension, or in other words, the number of spatial and temporal neighbours in optimal  
phase space reconstruction. We use the method of false neighbours \cite{1992PhRvA..45.3403K, 1992PhRvA..45.7058M,1993RvMP...65.1331A},
which determines that falsely apparent close neighbours have been 
eliminated by virtue of projecting the full orbit in a increasing higher dimensional embedding phase space. This gives us the 
$J^*$, the optimal number of time slices to take,
and $I^*$, the optimal number of spatial slices to take in our ${\bf x}(s^n_m)$ optimal reconstruction.

In this article, we propose that as any set of input representation 
``approaches'' the optimal one, then $\textnormal{SSIM} \to 1$. In the case of finite training sets and/or noisy training sets  
$\textnormal{SSIM} \to x<1$, where $x$ is the best forecast possible given the data set. Visually, we believe that the SSIM versus some reasonable 
metric constructed to represent the distance between any input representation and the optimal input representation will show a skewed bell shape 
as depicted in Fig.\ \ref{conjecture}. In this proposal, we use the most obvious candidate to represent the distance between any input 
representation and the optimal   input representation, the Euclidian distance given by   
$d_e=\sqrt{(I-I^*)^2+(J-J^*)^2+(K-K^*)^2+(L-L^*)^2}$, where $I$,$J$,$K$,$L$ are the parameters for each representation
and $I^*$,$J^*$,$K^*$,$L^*$ are the ones derived from the dynamical systems theory. We also verified that other reasonable metrics,   
in particular the Manhattan distance\cite{0486252027}, did 
not change the results qualitatively.

%\IEEEpubidadjcol

\section{Results}
\label{resultssection}

In order to empirically substantiate our proposal, we take four examples of spatial-temporal series and attempt
to forecast using our feedforward neural network. 
First, we split the data into a training and a test set. Second, using the training set only, 
we calculate the optimal time delay/spatial delays ($L^*$ and $K^*$, respectively)
using the first minima of the mutual information, and then we calculate the optimal number of temporal/spatial points to use
($J^*$ and $I^*$, respectively) using the method of false nearest neighbours.
Only then we build the neural network model, calibrating the hyperparameters of the network by exhaustive search on the parameter space
to minimize the error on the training set. Then having fixed those hyperparameters, we 
use a Monte Carlo simulation on the test set, on each one of our four examples, sampling random values
of the key feature selection parameters: $I$, $J$, $K$, $L$ (including the trivial ones with $I=0$ and/or $J=0$) 
and calculate the values $d_e(I,J,K,L)$ 
and $\textnormal{SSIM}(I,J,K,L)$.
We plot the latter as a function of the former to compare against our proposal as depicted in Fig.~\ref{conjecture}.

We first take a physical system example, a real data example, and then we progress from ``simpler'' systems (coupled maps) 
capable of generating 
spatial-temporal chaos to more ``complex'' systems (coupled Ordinary Differential Equations - ODEs) to ``really complex'' systems (Partial 
Differential Equations - PDEs). This is partially motivated by results in the literature that show that general universalities are 
present in different levels of simplification of physical models \cite{2001Chaos..11..404C}, 
from the original PDEs to truncated ODE expansions (e.g. spectral 
method expansions \cite{2001cfsm.book.....B}) to the most extreme simplification or discretization such as maps 
which capture the essence of the problem(s). In all cases we take examples with one spatial and one temporal dimension. 
However, we believe 
that our proposal will extend to multiple spatial dimensions. Again, notice that here we are not trying to demonstrate
that neural networks, and in particular feedforward neural networks can perform well in predicting spatial-temporal chaos
(as this has been demonstrated in the literature already), but 
rather to show that the optimal choice of the input layer features is given by dynamical systems theory and does not need to be
another neural network hyperparameter calculated by the ``dark art'' of trial and error.

\subsection{Sunspot data - a physical system example}

% code and results in
% C:\Users\eurico\SunspotAnalysis\NeuralNetworks\SolarForecastingNeuralNetworks41.xlsm

\begin{figure}[!htb]
\centering
\resizebox{\hsize}{!}{\includegraphics[]{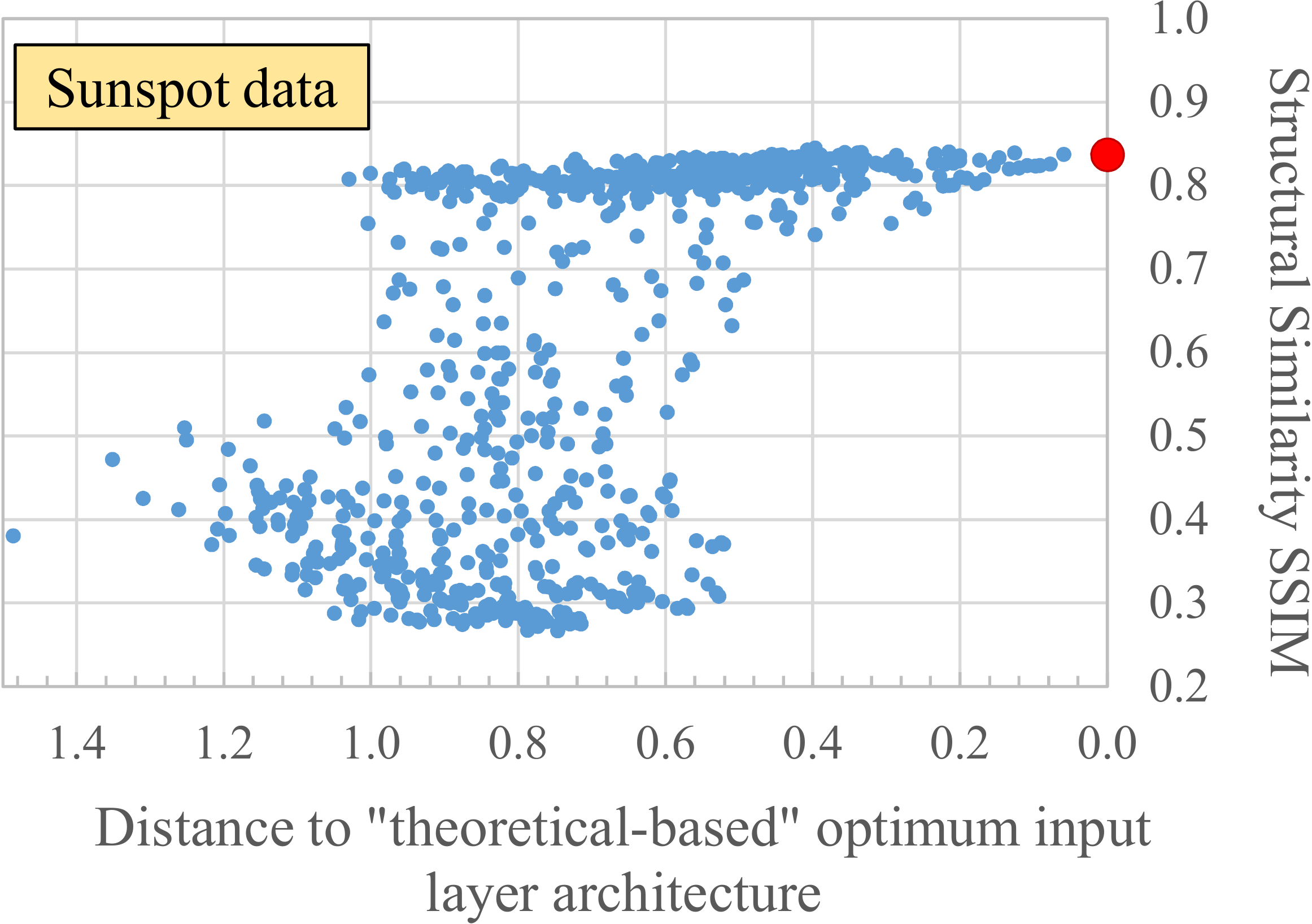}}
\caption{Monte Carlo simulation of different  input representations  of the input layer for the neural network forecast for the sunspot data.
It shows the structural similarity (SSIM) against how far (in a Euclidean space metric) the particular parameters of a particular
run were from the supposedly optimal input representation parameters (red dot).}
\label{MonteCarloSSIMversusParameterMetricDistance}
\end{figure}

The first example we take is a physical real data example based on a previous article of one of us \cite{covaspeixinhojoao}, where a 
neural network using the type of input representation above (Fig.\ \ref{architecture}) was used to forecast sunspot areas $A(t,\theta)$ in our 
Sun in both space ($\theta$ latitude) and time (Carrington Rotation index\footnote{Given that the surface solar rotation 
varies with time and latitude,
any approach of comparing positions on the Sun over a period of time is necessarily subjective.
Therefore, solar rotation is arbitrarily taken to be 27.2752316 days
for the purpose of Carrington rotations. Each solar rotation is given a 
number, the so-called Carrington Rotation Number, starting from 9th November, 1853.}). This sunspot data is usually called the
``butterfly diagram'' due to its butterfly wings like appearance \cite{1904MNRAS..64..747M}. One can see how this butterfly diagram looks like in
\cite{BibEntry2018Mar}.

Sunspot data is regularly seen as a benchmark for time series forecasting, given its chaotic nature and that it considered to be among the 
longest continuously recorded daily measurement made in science \cite{2013Natur.495..300O}.
Many authors \cite{
1990EOSTr..71..677K,
1991PhDT.......158W,
Weigend92HubermanRumelhart,
1993AdSpR..13..447M,
1995JGR...10021735M,
1994VA.....38..351C,
0305-4470-28-12-012,
1995ApJ...444..916C,
Koskela96timeseries,
1996SoPh..168..423F,
1996AnGeo..14...20F,
1996ITNN....7..501P,
1998GeoRL..25..457K,
1998JGR...10329733C,
1998NewAR..42..343C,
Verdes2000,
2004SoPh..221..167V,
2001GMS...125..201L,
2002PhRvE..66f6701S,
2004SoPh..224..247M,
2005JASTP..67..595G,
2004SPIE.5497..542A ,
2005MmSAI..76.1030Q,
2005SoPh..227..177A,
2006JASTP..68.2061M,
2007SoPh..243..253Q ,
2013Ap&SS.344....5A,
xie2006hybrid,
2006SunGe...1a...8M,
2007IJMPC..18.1839E,
1997SPIE.3077..116P,
2006AGUFMSH21A0315L,
2008cosp...37.3467W,
gang2007sunspot,
2009JASTP..71..569U,
2010BAAA...53..241F,
1999BAAA...43...23P,
2011RAA....11..491A,
2010cosp...38.2153A,
2011CRGeo.343..433C,
JiangS11,
2012cosp...39.1194M,
Chandra:2012:CCE:2181341.2181747,
park2009prediction ,
kim2010sunspot,
moghaddam2013sunspot,
2012EPJP..127...43C,
liu2012sunspot,
Gkana201579,
DBLP:conf/ijcnn/ParsapoorBS15,
DBLP:conf/aaai/ParsapoorBS15,
raios
} have already attempted to use neural networks to forecast aspects of the sunspot cycle, although as far as we are aware, 
none in both space and time having 
restricted themselves to using these neural networks to forecast mostly either the sunspot number or the sunspot areas as a function of 
time.  There is only one example\cite{covaspeixinhojoao}, as far as we are aware, of actual spatial-temporal forecasts using neural networks (see also 
\cite{2006AGUFMSH21A0315L,2008cosp...37.3467W} where a neural network forecast of the magnetic flux, which is related to sunspots, is forecast
for latitude/longitude datasets). There are also a few examples of forecasting the butterfly diagram sunspot data in both space and time (latitude/time)
\cite{2011A&A...528A..82J, 2016ApJ...823L..22C, 0004-637X-792-1-12, 2017arXiv170700268J, covas2016, 2017arXiv171207501S} but none of these used neural networks, rather
all of those used other statistical methods or numerical physical modelling.

We take as a ``training set'' the data from the year 1874 to 
approximately 1997 (i.e.\ the first 1646 Carrington Rotations). We then attempt to reproduce or forecast the sunspot area butterfly 
diagram from Carrington Rotation 1921 up to 2162 (the last one corresponding approximately to the year 2015); that is, we use 1646 time 
slices ($\approx 122.92$ years) to reproduce the next 242 time slices ($\approx 18.07$ years)\footnote{We use exactly
the same training set/forecast set slicing as in \cite{covas2016,covaspeixinhojoao} for consistency, even if more data is already
available at this time.}. The training set 
corresponds to around 12 solar cycles (cycle 11 to 22), while the ``forecasting set'' equates to around 1.5 cycles 
(cycle 23 and half of cycle 24). The entire dataset, including the training and forecasting sets, is a grid $x^i_j=x(i,j)$, with 
$i=1888$ and $j=50$. The training set is a grid $x(1646,50)$. For this case the optimal values were $I^*=2$, $J^*=6$, $K^*=9$ and 
$L^*=70$ as calculated in \cite{covas2016}. 
The hyperparameters of the neural network were:
$N_h=70$, $\eta=0.3$, $\alpha=0.01$, a logarithmic normalization of the inputs scaled
 with $\alpha_{nor} = 10$ and $\beta_{nor} = 0$, weight initialization with $\alpha_{rng} = 10^{-2}$ and $\beta_{rng} = -0.5$ 
 and $N_{\textnormal{steps}}=$\SI{1000000}\nobreak. We used the logistic sigmoid function as the activation on both the hidden and output layers.

The Monte Carlo results are depicted in Fig.\ \ref{MonteCarloSSIMversusParameterMetricDistance} showing runs with different $I$, $J$, $K$, $L$
and plotting the SSIM versus the distance to the optimal input feature selection parameters ($I^*$,$J^*$,$K^*$,$L^*$) 
given by the dynamical systems theory.  
It shows a reasonable expected dispersion as proposed and 
a good convergence to the highest $\textnormal{SSIM}$ value we could obtain for this particular slicing of the training and forecast sets 
$\textnormal{SSIM}= 0.836876152$. From the figure, there seems to be also two clusters of behaviour, and at closer inspection, we found that the
cluster with lower SSIM is basically a set of very bad forecasts, with none of the characteristics of the real sunspot behaviour (the 11 year-like cycle
and the migration to the latitudinal equator), while the higher SSIM cluster corresponds to visually recognizable sunspot butterfly-like diagrams.

These results were quite satisfactory and inspired us to attempt to check the existence of a universality of behaviour across
dynamical systems, by examining other unrelated synthetic
generated data sets. We continue below to these attempts.

% An example of a floating figure using the graphicx package.
% Note that \label must occur AFTER (or within) \caption.
% For figures, \caption should occur after the \includegraphics.
% Note that IEEEtran v1.7 and later has special internal code that
% is designed to preserve the operation of \label within \caption
% even when the captionsoff option is in effect. However, because
% of issues like this, it may be the safest practice to put all your
% \label just after \caption rather than within \caption{}.
%
% Reminder: the "draftcls" or "draftclsnofoot", not "draft", class
% option should be used if it is desired that the figures are to be
% displayed while in draft mode.
%

\subsection{Coupled H\'{e}non maps - a discrete-time dynamical system}

\begin{figure}[!htb]
\centering
\resizebox{\hsize}{!}{\includegraphics[]{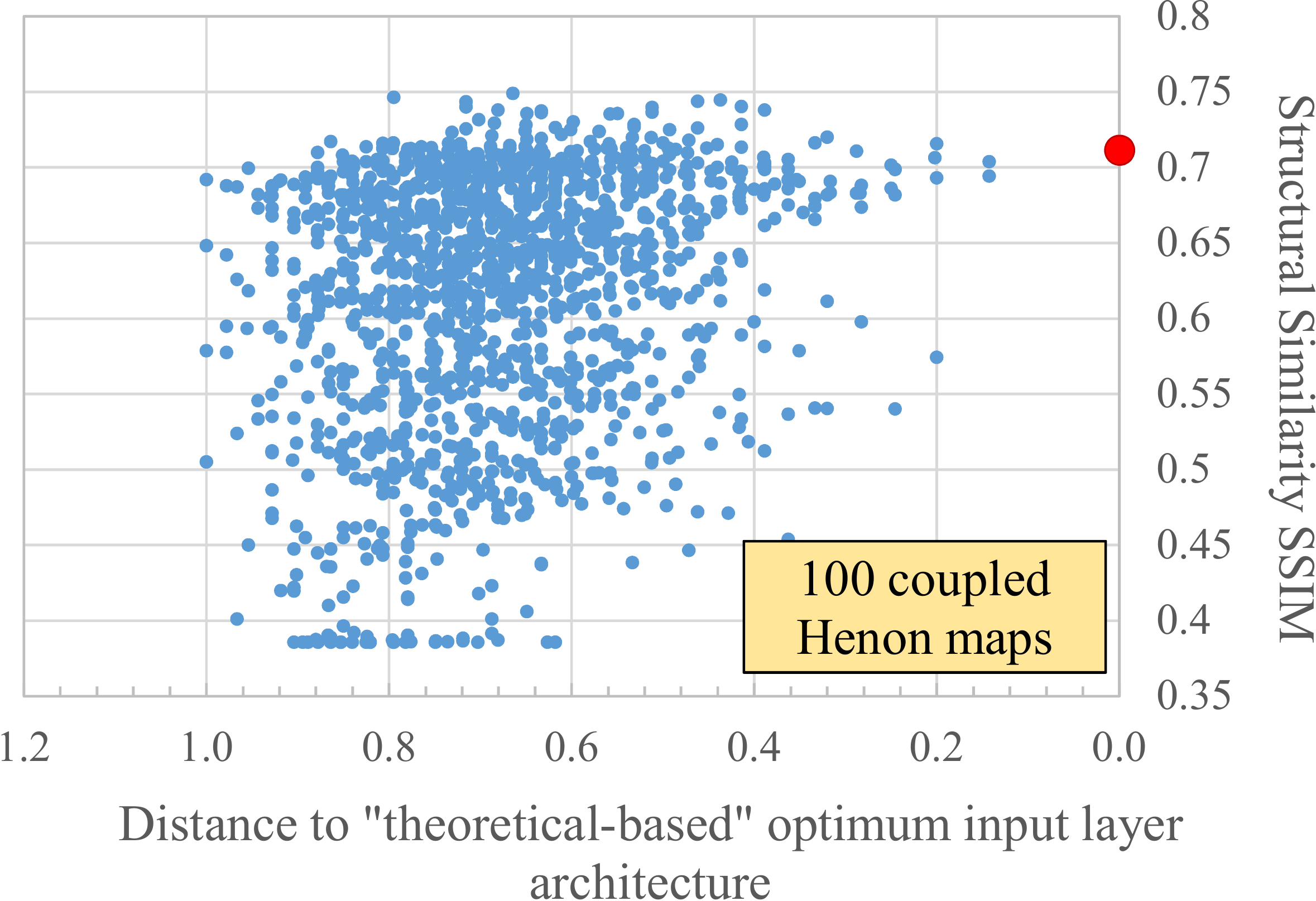}}
\caption{Monte Carlo simulation of different  input representations of the input layer for the neural network forecast for a series of 100 coupled
H\'{e}non maps.
It shows the structural similarity (SSIM) against how far (in a Euclidean space metric) the particular parameters of a particular
run was from the supposedly optimal input representation parameters (red dot). The green line (trendline) seems to show that as the parameters
of a randomly chosen input representation get close to the supposedly optimal input representation ones, the SSIM converges to what seems to be the
best possible forecast value given the limited dataset.}
\label{MonteCarloSSIMversusParameterMetricDistance100HenonCoupledMaps}
\end{figure}

% code and results in
% C:\Users\eurico\SunspotAnalysis\SpatialTemporalFeatureSelectionOptimalApproach\KuramotoSivashinsky27_HenonMaps.xlsm

Motivated by having a real case from a physical system, we then tried to investigate if this same proposal
holds in a very simplified example of a spatial-temporal model. Coupled maps are widely used as models of spatial-temporal
chaos and pattern/structure formation \cite{1989PThPS..99..263K,1989JSP....54.1489M,9780471937418}.
Following \cite{2000PhRvL..84.1890P,Parlitz2000NonlinearPO} we take a lattice of $M=100$
coupled H\'{e}non maps:
\begin{IEEEeqnarray}{lCr}
\label{henon}
u_m^{n+1} = 1 - 1.45 \left[ \frac{1}{2} u^n_m + \frac{u_{m-1}^{n}+u_{m+1}^{n}}{4} \right]^2 + 0.3 v_m^n, &&\\
v_m^{n+1} = u_m^n.&& \nonumber
\end{IEEEeqnarray}
with fixed boundary conditions $u^n_1=u_M^n=\frac{1}{2}$ and $v_1^n=v_M^n=0$. The initial values for rest of the variables 
$u^{n=0}_{m\neq 1,M}$
and $v^{n=0}_{m\neq 1,M}$ is taken from a random constant distribution in the range $[0,1[$.

We run the synthetic data generation for $N=531$ time steps, and divided the set into $N_{\textnormal{train}}=500$ time steps for the training set
and $N_{\textnormal{test}}=31$ time steps for the test set. The other parameters of the neural network were:
$N_h=10$, $\eta=0.1$, $\alpha=0$, a linear input normalization scaling with $\alpha_{nor} = 2.947992$, 
$\beta_{nor} = 0.515$, $\alpha_{rng} = 10^{-3}$, $\beta_{rng} = -0.5$ and $N_{\textnormal{steps}}=$\SI{1000000}\nobreak. 
We used the ReLu function as the activation on both the hidden and output layers.

For this case the optimal values given by the mutual information and the false neighbours methods were $I^*=1$, $J^*=3$, $K^*=2$ and $L^*=3$. 
The results of the Monte Carlo simulation for different $I$, $J$, $K$ and $L$
are depicted in Fig.\ \ref{MonteCarloSSIMversusParameterMetricDistance}.  It again shows a dispersion as proposed and a reasonable
convergence to the highest
 $\textnormal{SSIM}$ value we could obtain for this particular slicing of the training and forecast sets $\textnormal{SSIM}=
0.71139101$.

Results suggest the same structure as depicted in our proposal diagram and in the previous results for sunspots. We now move
below to a more complex model, a coupled set of ODEs.

\subsection{Coupled Ordinary Differential Equations - Lorenz-96 model}

\begin{figure}[!htb]
\centering
\resizebox{\hsize}{!}{\includegraphics[]{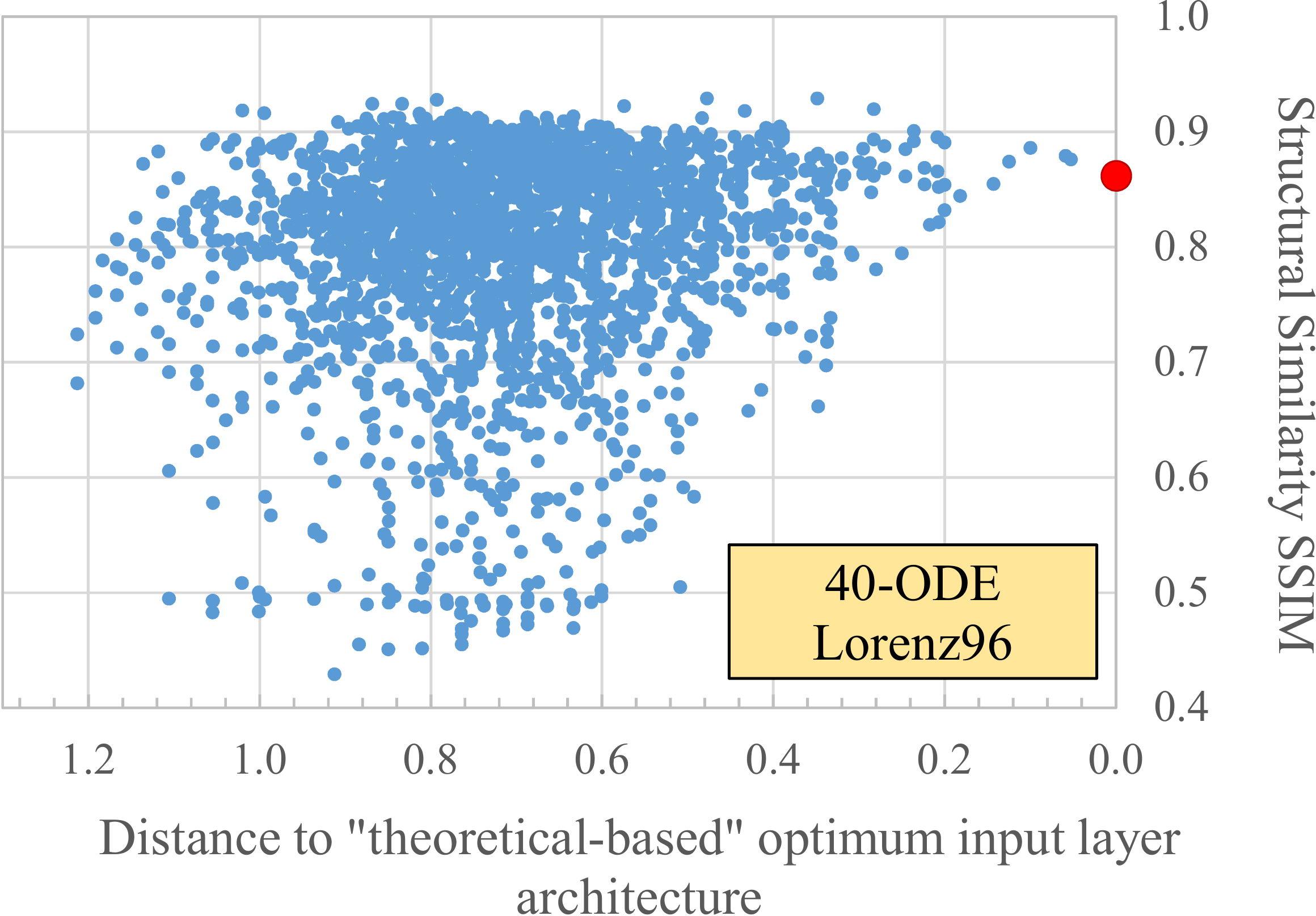}}
\caption{Monte Carlo simulation of different  input representations of the input layer for the neural network forecast for the 40-ODE Lorenz 96 system.
It shows the structural similarity (SSIM) against how far (in a Euclidean space metric) the particular parameters of a particular
run was from the supposedly optimal input representation parameters (red dot). The green line (trendline) seems to show that as the parameters
of a randomly chosen input representation get close to the supposedly optimal input representation ones, the SSIM converges to what seems to be the
best possible forecast value given the limited (and noisy) dataset.}
% https://en.wikipedia.org/wiki/Lorenz_96_model
\label{MonteCarloSSIMversusParameterMetricDistanceLorenz96}
\end{figure}

% code in
% C:\Users\eurico\SunspotAnalysis\SpatialTemporalFeatureSelectionOptimalApproach\lorenz4D\lorenz4Drun.m
%%% the Lorenz model is: (cyclical)
% dX[j]/dt=(X[j+1]-X[j-2])*X[j-1]-X[j]+F
%J=40;               %the number of variables
%h=0.05;             %the time step
% results in
% C:\Users\eurico\SunspotAnalysis\SpatialTemporalFeatureSelectionOptimalApproach\KuramotoSivashinsky23_Lorenz.xlsm

For the spatially extended coupled ODEs model we used a well-known 40-coupled ODE dynamical system proposed by Edward Lorenz in 1996 
\cite{articleLorenz96}:

\begin{equation}
\label{lorenz96equations}
\frac{dx_j}{dt}=\left( x_{j+1} - x_{j-2} \right ) x_{j-1} - x_j + F, \, j=1,\ldots,N=40,
\end{equation}
where $x_{-1}=x_{N-1}$, $x_0=x_N$ and $x_{N+1}=x_1$ and $F$ is a forcing term. 
We use the forcing $F=5$ to get some interesting behaviour in space and time. We used a time step $\Delta t=0.05$ and 
we have integrated this equation using J.\ Amezcua's MATLAB code as given in \cite{BibEntry2018Apr}. It uses the 
Runge-Kutta 4-step method.

We run the synthetic data generation for $N=531$ time steps, and divided the set into $N_{\textnormal{train}}=500$ time steps for the training set
and $N_{\textnormal{test}}=31$ time steps for the test set. The other parameters of the neural network were:
$N_h=10$, $\eta=0.05$, $\alpha=0.001$, a linear normalization input scaling with $\alpha_{nor} = 10$ and $\beta_{nor} = 0.430$, 
weight initialization with $\alpha_{rng} = 10^{-3}$ and $\beta_{rng} = -0.5$ and $N_{\textnormal{steps}}=100,000$. 
We used the ReLU function as the activation on both the hidden and output layers.

For this case the optimal values obtained before the Monte Carlo simulation from the mutual information and false neighbours methods 
were $I^*=2$, $J^*=2$, $K^*=1$ and $L^*=9$. The results of the random sampling of $I$, $J$, $K$, $L$ in the simulation
are depicted in Fig.\ \ref{MonteCarloSSIMversusParameterMetricDistanceLorenz96}.  It shows a dispersion as proposed and a quite a good
convergence to the highest 
$\textnormal{SSIM}$ value we could obtain for this particular slicing of the training and forecast sets: $\textnormal{SSIM}=
0.861844038$. Results suggest the same structure as depicted in our proposal diagram and in the previous results for sunspots
and the coupled H\'{e}non maps.

\subsection{Partial Differential Equations - Kuramoto-Sivashinsky model}

\begin{figure}[!htb]
\centering
\resizebox{\hsize}{!}{\includegraphics[]{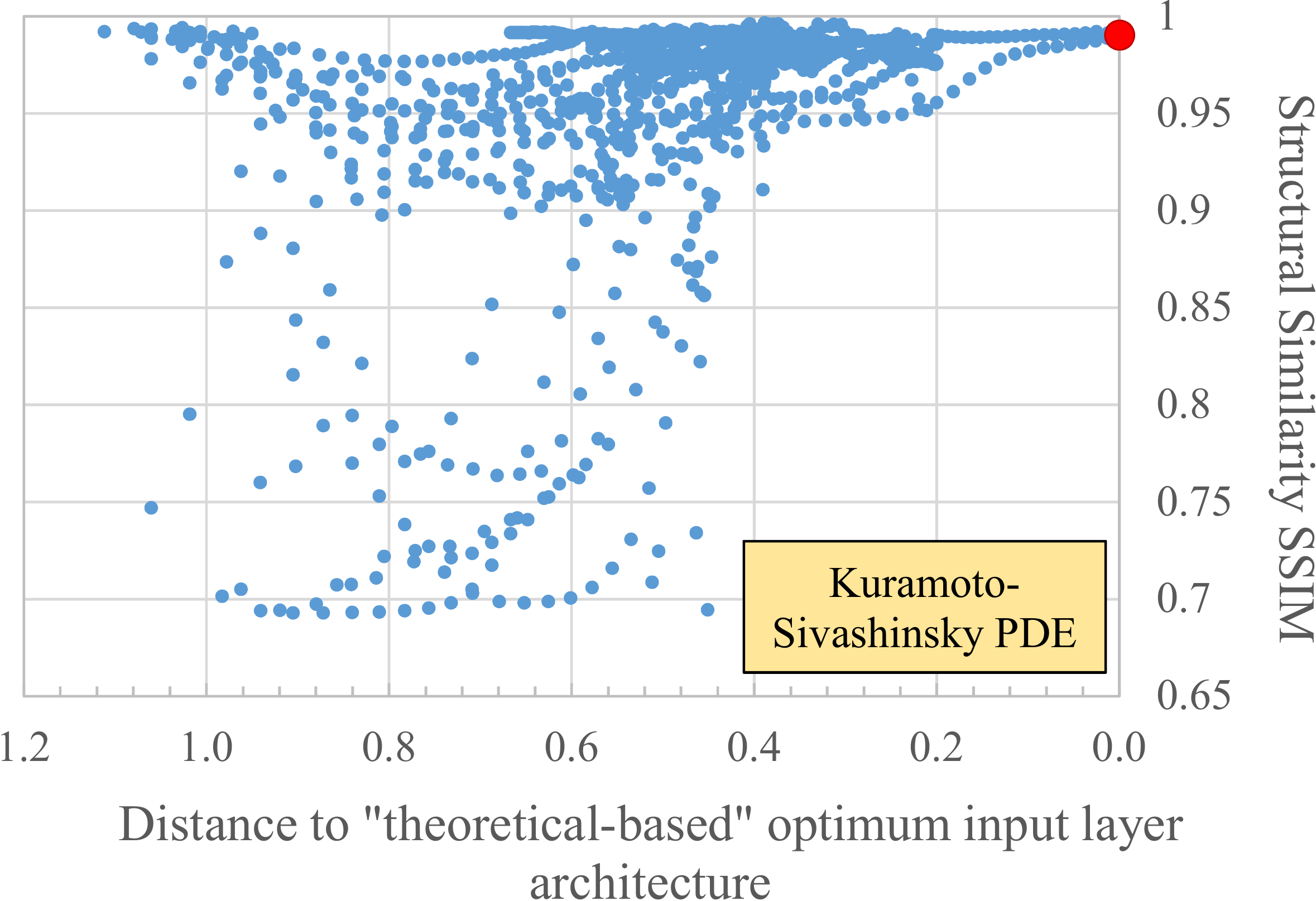}}
\caption{Monte Carlo simulation of different input representations of the input layer for the neural network forecast for
Kuramoto-Sivashinsky with $L=22$ system.
It shows the structural similarity (SSIM) against how far (in a Euclidean space metric) the particular parameters of a particular
run was from the supposedly optimal input representation parameters (red dot). The green line (trendline) seems to show that as the parameters
of a randomly chosen input representation get close to the supposedly optimal input representation ones, the SSIM converges to what seems to be the
best possible forecast value given the limited (and noisy) dataset.}
\label{MonteCarloSSIMversusParameterMetricDistanceKS_L=22}
\end{figure}

% code in
% C:\Users\eurico\SunspotAnalysis\SpatialTemporalFeatureSelectionOptimalApproach\KSEproject\matlabKSintrinsicSolver.m
% from http://chaosbook.org/extras/KSEproject/html/index.html
% results in
% C:\Users\eurico\SunspotAnalysis\SpatialTemporalFeatureSelectionOptimalApproach\KuramotoSivashinsky28.xlsm

Finally we take a full PDE system, the Kuramoto-Sivashinsky model \cite{1976PThPh..55..356K,1977AcAau...4.1177S},
a very well-known system capable of spatial-temporal chaos
and complex spatial-temporal dynamics. It is a fourth-order nonlinear PDE introduced in the 1970s by Yoshiki
Kuramoto and Gregory Sivashinsky to model the diffusive instabilities
in a laminar flame front.
The model is described by the following equation:

\begin{equation}
\label{kuramotoSivashinskyequation}
\frac{\partial u(x,t)}{\partial t} = -\frac{\partial^4 u(x,t)}{\partial x^4}-\frac{\partial^2 u(x,t)}{\partial x^2}-u(x,t)
\frac{\partial u(x,t)}{\partial x},
\end{equation}
where $x \in [-\frac{L}{2},+\frac{L}{2}]$ with a period boundary condition 
$u(x+L,t)=u(x,t)$. The nature of solutions depends on the system size $L$ and on the initial $u(x,t=0)$. 
We have integrated this equation by taking an exponential time difference Runge-Kutta 4th order method (ETDRK4)
using the Matlab code by P.\ Cvitanovi\'c as given in \cite{BibEntry2007Apr} and 
taking a time step of $\Delta t=0.5$, $L=22$ Fourier modes which are known to produce a ``turbulent'' or chaotic behaviour
and a initial condition $u(x,t=0)=10^{-5}$ for $x\in[5,15]$, the remain being $u(x,t=0)=0$.

We run the simulation for $N=531$ time steps, and divided the set into $N_{\textnormal{train}}=500$ time steps for the training set
and $N_{\textnormal{test}}=31$ time steps for the test set. The other parameters of the neural network were:
$N_h=50$, $\eta=0.1$, $\alpha=0$, a linear normalization input scaling with 
$\alpha_{nor} = 5.8472$ and $\beta_{nor} =0.5$, weight initialization 
with $\alpha_{rng} = 10^{-3}$ and $\beta_{rng} = -0.5$ and $N_{\textnormal{steps}}=$\SI{1000000}\nobreak. 
We used the ReLU function as the activation on both the hidden and output layers.

The results of the Monte Carlo simulation can be seen in Fig.\ \ref{MonteCarloSSIMversusParameterMetricDistanceKS_L=22}. For this 
case the optimal values obtained before we run the Monte Carlo simulation were $I^*=1$, $J^*=2$, $K^*=2$ and $L^*=39$. It again shows a dispersion as 
proposed and a excellent convergence to the highest $\textnormal{SSIM}$ value we could obtain for this particular slicing of the training and 
test sets: a surprising high value of $\textnormal{SSIM}=0.990264382$. Results suggest the same structure as depicted in our 
proposal diagram and in the previous results for sunspots, the coupled H\'{e}non maps and coupled ODEs.

\section{Conclusions}
\label{conclusionsection}
In this paper, we have shown empirical evidence for the existence of an optimal feature selection for the input layer of feedforward neural networks used to 
forecast spatial-temporal series. We believe that the selection of the features of the input layer can be uniquely determined by 
the data itself, using two techniques from dynamical systems embedding theory: the mutual information and the false neighbours 
methods. The former procedure determines the temporal and spatial delays to take when selecting features, while the latter  
determines the number of data points in space and time to be taken as inputs.  We propose that this optimal feature selection gives the best 
forecast, as measured by a standard image similarity index. We also propose that the shape of the dispersion of points on a Monte Carlo 
simulation across all possible feature selections on a plot of the similarity index versus the 
distance to optimal feature selection is a skewed bell  
shape with the highest value being the optimal feature selection/maximum similarity index.
  
In order to substantiate our proposal, we chose four unrelated systems, in order of complexity: a set of spatially 
extended coupled maps; a set of spatially extended coupled ODEs; a one-dimensional spatial PDE and a real spatial-temporal data set from 
sunspots areas in our Sun. In all four cases, we were able to first use the mutual information and the false neighbours methods to 
determine the four parameters defining the input layer feature selection\footnote{We have four parameters for the feature selection 
in these cases, with one temporal and one spatial dimension. For higher dimensional systems, there will be more parameters, the exact 
number being double the number of dimensions of the system.}. After calibration of the hyperparameters we then were able to forecast 
reasonably the test set, although this is not the objective or primary goal of this article. 
We then show that for a random Monte Carlo simulation across possible feature selections, the neural network did not, 
as expected, forecast as well as it did for the specific set of optimal four parameters given by dynamical systems theory. 
As proposed, the Monte Carlo 
simulations show that the shape of the distribution of points was a 
skewed bell shape with the highest value being the optimal feature selection/maximum similarity index (subject to minor variations due to noise
and the finiteness of the dataset).

Given how important spatial-temporal systems are 
and how we want to forecast the future as accurately as possible 
it is quite 
important to attempt to reduce the number of hyperparameters in neural network prediction, and to try to constrain the feature selection from 
the data properties only. If indeed our proposal turns out to be true, it would remove the input layer feature selection as another free 
parameter in the already complex process of choosing the details of the neural network to use for forecasting.

In this article we have focused first and foremost in establishing empirical evidence for our proposal, within a simple framework of 
feedforward neural networks with one hidden layer for the purpose of prediction in one spatial and one temporal dimensions. Naturally, 
there are many clear extensions to our research. First to use deeper networks with additional hidden 
layers to possibly tackle systems which are hyperchaotic (i.e. with multiple positive Lyapunov exponents). Second, to attempt to extend the proposal with empirical evidence in high 
dimensions, e.g. 3+1-dimensional weather systems. Third, to extend the proposal to other commonly used 
neural network models, such as recurrent neural networks\cite{COGS:COGS203}, particularly echo state networks \cite{2015arXiv151007146M,2017arXiv170805094M} 
and long short-term memory networks \cite{Hochreiter:1997:LSM:1246443.1246450}.
Fourth and last but not least, to demonstrate the proposal rigorously would show how dynamical systems theory can clarify the so called
``dark art'' in neural network feature construction. These objectives are however, outside the scope
of this research article and will be pursued as part of future work.

% or


%\appendices

%\section{}

\section*{Acknowledgments}
We would like to thank Prof. Reza Tavakol from Queen Mary University of London for very useful discussions on forecasting. We also thank
Dr.\ David Hathaway from NASA's Ames Research Centre for providing the sunspot data on which some of the results in this article are based upon.
CITEUC is funded by National Funds through FCT - Foundation for Science
and Technology (project: UID/Multi/00611/2013) and FEDER - European
Regional Development Fund through
COMPETE 2020 - Operational Programme Competitiveness and
Internationalization (project: POCI-01-0145-FEDER-006922).
EB is supported by a UK RAEng Research Fellowship (RF/128).

\ifCLASSOPTIONcaptionsoff
  \newpage
\fi



\newcommand*\aap{A\&A}
\let\astap=\aap
\newcommand*\aapr{A\&A~Rev.}
\newcommand*\aaps{A\&AS}
\newcommand*\actaa{Acta Astron.}
\newcommand*\aj{AJ}
\newcommand*\ao{Appl.~Opt.}
\let\applopt\ao
\newcommand*\apj{ApJ}
\newcommand*\apjl{ApJ}
\let\apjlett\apjl
\newcommand*\apjs{ApJS}
\let\apjsupp\apjs
\newcommand*\aplett{Astrophys.~Lett.}
\newcommand*\apspr{Astrophys.~Space~Phys.~Res.}
\newcommand*\apss{Ap\&SS}
\newcommand*\araa{ARA\&A}
\newcommand*\azh{AZh}
\newcommand*\baas{BAAS}
\newcommand*\bac{Bull. astr. Inst. Czechosl.}
\newcommand*\bain{Bull.~Astron.~Inst.~Netherlands}
\newcommand*\caa{Chinese Astron. Astrophys.}
\newcommand*\cjaa{Chinese J. Astron. Astrophys.}
\newcommand*\fcp{Fund.~Cosmic~Phys.}
\newcommand*\gca{Geochim.~Cosmochim.~Acta}
\newcommand*\grl{Geophys.~Res.~Lett.}
\newcommand*\iaucirc{IAU~Circ.}
\newcommand*\icarus{Icarus}
\newcommand*\jcap{J. Cosmology Astropart. Phys.}
\newcommand*\jcp{J.~Chem.~Phys.}
\newcommand*\jgr{J.~Geophys.~Res.}
\newcommand*\jqsrt{J.~Quant.~Spec.~Radiat.~Transf.}
\newcommand*\jrasc{JRASC}
\newcommand*\memras{MmRAS}
\newcommand*\memsai{Mem.~Soc.~Astron.~Italiana}
\newcommand*\mnras{MNRAS}
\newcommand*\na{New A}
\newcommand*\nar{New A Rev.}
\newcommand*\nat{Nature}
\newcommand*\nphysa{Nucl.~Phys.~A}
\newcommand*\pasa{PASA}
\newcommand*\pasj{PASJ}
\newcommand*\pasp{PASP}
\newcommand*\physrep{Phys.~Rep.}
\newcommand*\physscr{Phys.~Scr}
\newcommand*\planss{Planet.~Space~Sci.}
\newcommand*\pra{Phys.~Rev.~A}
\newcommand*\prb{Phys.~Rev.~B}
\newcommand*\prc{Phys.~Rev.~C}
\newcommand*\prd{Phys.~Rev.~D}
\newcommand*\pre{Phys.~Rev.~E}
\newcommand*\prl{Phys.~Rev.~Lett.}
\newcommand*\procspie{Proc.~SPIE}
\newcommand*\qjras{QJRAS}
\newcommand*\rmxaa{Rev. Mexicana Astron. Astrofis.}
\newcommand*\skytel{S\&T}
\newcommand*\solphys{Sol.~Phys.}
\newcommand*\sovast{Soviet~Ast.}
\newcommand*\ssr{Space~Sci.~Rev.}
\newcommand*\zap{ZAp}

%\bibliographystyle{IEEEtran}
\bibliographystyle{IEEEtran}
%\bibliography{IEEEabrv,eurico} % your references Yourfile.bib
\bibliography{eurico} % your references Yourfile.bib
\end{document}